\documentclass[12pt,reqno]{amsart}

\usepackage{amsmath,amssymb,amsthm}
\usepackage{amsaddr}
\usepackage{bm}
\usepackage[dvipdfmx]{graphicx} 
\usepackage{here,comment}
\usepackage{algorithm,algorithmic}
\usepackage{xcolor}
\usepackage{fullpage}
\usepackage{booktabs}

\theoremstyle{definition}

\usepackage{comment}

\renewcommand{\hat}{\widehat}

\newcommand{\argmin}{\operatornamewithlimits{argmin}}

\usepackage{color}

\usepackage{newtxtext,newtxmath}

\title{Effect of Weight Quantization on Learning Models \\by Typical Case Analysis}

\author{Shuhei Kashiwamura$^1$, Ayaka Sakata$^2$, Masaaki Imaizumi$^{1,3}$}
\address{The University of Tokyo$^1$, The Institute of Statistical Mathematics$^2$, RIKEN AIP$^3$}

\date{\today}

\allowdisplaybreaks

\begin{document}
\maketitle

\begin{abstract}
   This paper examines the quantization methods used in large-scale data analysis models and their hyperparameter choices. The recent surge in data analysis scale has significantly increased computational resource requirements. To address this, quantizing model weights has become a prevalent practice in data analysis applications such as deep learning. Quantization is particularly vital for deploying large models on devices with limited computational resources. However, the selection of quantization hyperparameters, like the number of bits and value range for weight quantization, remains an underexplored area. In this study, we employ the typical case analysis from statistical physics, specifically the replica method, to explore the impact of hyperparameters on the quantization of simple learning models. Our analysis yields three key findings: (i) an unstable hyperparameter phase, known as replica symmetry breaking, occurs with a small number of bits and a large quantization width; (ii) there is an optimal quantization width that minimizes error; and (iii) quantization delays the onset of overparameterization, helping to mitigate overfitting as indicated by the double descent phenomenon. We also discover that non-uniform quantization can enhance stability. Additionally, we develop an approximate message-passing algorithm to validate our theoretical results.
\end{abstract}

\section{Introduction}
The development of data-driven science, which involves analyzing large-scale observational data using models with a vast number of parameters, has led to a significant increase in computational costs.
A prime example is the large-scale neural networks used in artificial intelligence technologies.
For instance, GPT (Generative Pre-trained Transformer)-3 \cite{brown2020language} possesses tens of billions of parameters, requiring substantial computational resources for storage and updates in computer memory.
Moreover, with the growing need to implement these models in small-edge devices, it's crucial to minimize their computational resource requirements.
Practical applications, such as implementing neural networks in smartphones and in-vehicle sensors, are examples of this trend \cite{liang2021pruning}.
Reducing computational costs is essential to achieve accurate and sustainable inference on these devices.

\textit{Quantization} is one of the most common techniques for compressing large models (for a survey, see \cite{gholami2022survey}). Quantization involves rounding the parameters or activation values of a model to discrete values with lower bit precision. Specifically, it converts original values stored with higher bits, e.g., 32 bits, to lower bits, such as 4 or 8 bits. This operation not only reduces the amount of memory required to store parameters but also decreases the computational resources needed for operations like matrix products. In practice, it is empirically known that quantization minimally impacts the prediction accuracy of neural networks, thereby enhancing the utility of this technology (see \cite{banner2018scalable} for an example). Additionally, quantization can be integrated with other model compression techniques, such as distillation and pruning \cite{liang2021pruning}. Its significance is anticipated to grow as models, including new versions of GPT, continue to increase in size.

An ongoing challenge in quantization is the selection of hyperparameters, including the number of bits and the quantization range. While neural networks exhibit some robustness to quantization, a significant reduction in the number of bits can make them susceptible to outliers and data shifts. Consequently, it is crucial to choose appropriate hyperparameters that balance accuracy with computational load. Nevertheless, the comprehension of quantization remains an evolving issue. The characteristics of optimal hyperparameters and their very existence are still not fully understood.

In this study, we investigate the effects of hyperparameters on quantization, namely the number of bits and the quantization range, and their impact on its generalization performance. To achieve this, we employ typical case analysis from statistical physics, the replica method, to analyze the characteristics of a simple learning model. Specifically, we focus on the learning model in the proportionally high-dimensional regime, where the number of parameters infinitely increases while maintaining a constant ratio to the sample size. Through this method, we derive the stability condition for the replica symmetric ansatz, which can be a stability measure for algorithms under quantization, detecting exponential number of local minima \cite{ricci2001simplest}, and subsequently ascertain the precise value of the generalization error in the high-dimensional regime. Additionally, we explore two quantization patterns, uniform and non-uniform quantization.

As for the theoretical contributions, this study elucidated the following points:
\begin{itemize}
    \item When the number of bits is small and the quantization range is broad, unstable phases intermittently appear, leading to algorithmic instability. Moreover, these unstable phases can be mitigated through non-uniform quantization.
    \item Regarding the selection of hyperparameters, greater bit numbers result in higher accuracy. Conversely, there exists an optimal quantization range that is neither excessively large nor too small.
    \item Quantization alters the peak of the double descent in the generalization error. In simpler terms, quantization slightly hinders reaching a phase where the learning model mitigates overfitting.
\end{itemize}
Additionally, we applied Approximate Message Passing (AMP) algorithm to the quantization that corresponds to the replica analysis. The simulation results from this algorithm validate our theoretical findings.

\subsection{Related Studies}
Papers on quantization, especially for neural networks, are numerous and have a long history. 
Only a few are mentioned here; the techniques of quantization prior to the 2000s are summarized in the paper \cite{gray1998quantization}. 
A seminal study \cite{banner2018scalable} demonstrated the utility of quantization in more modern deep learning and discussed the effectiveness of 8-bit quantization. This policy and the important role of bits in deep learning were actively discussed in \cite{wang2018training,wu2020integer,courbariaux2014training,baskin2021uniq}. 
More practical use of the quantization is also investigated: a study \cite{chmiel2020neural} demonstrated the effectiveness of 6-bit quantization techniques in improving performance with natural gradient methods. 
These research streams of quantization are exhaustively summarized in \cite{gholami2022survey}.


The theoretical study of quantization is still a developing field. 
This study \cite{ding2019universal} investigated the universal approximability of neural network models with the quantization, i.e., any continuous function can be approximated under any accuracy, as well as the magnitude of overfitting when making predictions with the quantized neural network. 
Another paper \cite{hernandez2023training} discussed theoretically the impact of the training process with the quantization on this prediction performance. 
This paper \cite{gonon2023approximation} quantified the impact of quantization on the approximation ability of neural networks. 
This study \cite{li2017training} analyzed theoretically the difficulty of learning by the quantization, and another study \cite{choi2016towards}  proposed a new loss function and methodology to prevent performance degradation due to the quantization.


Quantization has been a subject of investigation in statistical physics, particularly in classical spin models like the Ising model with binary variables and the Potts model with $K$-state values.
The variations in phase transition phenomena arising from 
distinct values of variables have been discussed \cite{Mezard1987}.
Statistical physics models and methods have been incorporated into statistical learning, 
and have successfully revealed conditions for both the failure and success of inference, as well as the development of the efficient algorithms \cite{kabashima2009typical,Krzakala2012PRX,Krzakala2012JSTAT,Zdeborova2016}.
In line with recent trends,
quantization settings are 
incorporated in statistical physics-based
machine learning studies,
such as 
Hopfield model with discrete coupling
\cite{sasaki2021analysis},
restricted Boltzmann machine with binary synapses \cite{huang2017statistical},
and single layer neural network with multiple states \cite{baldassi2016}.
However, there is still a lack of discussion on proper quantization, especially in terms of generalization, which is focus of our study.


\section{Problem Setup}
\begin{figure}
    \centering
    \includegraphics[width=150mm]{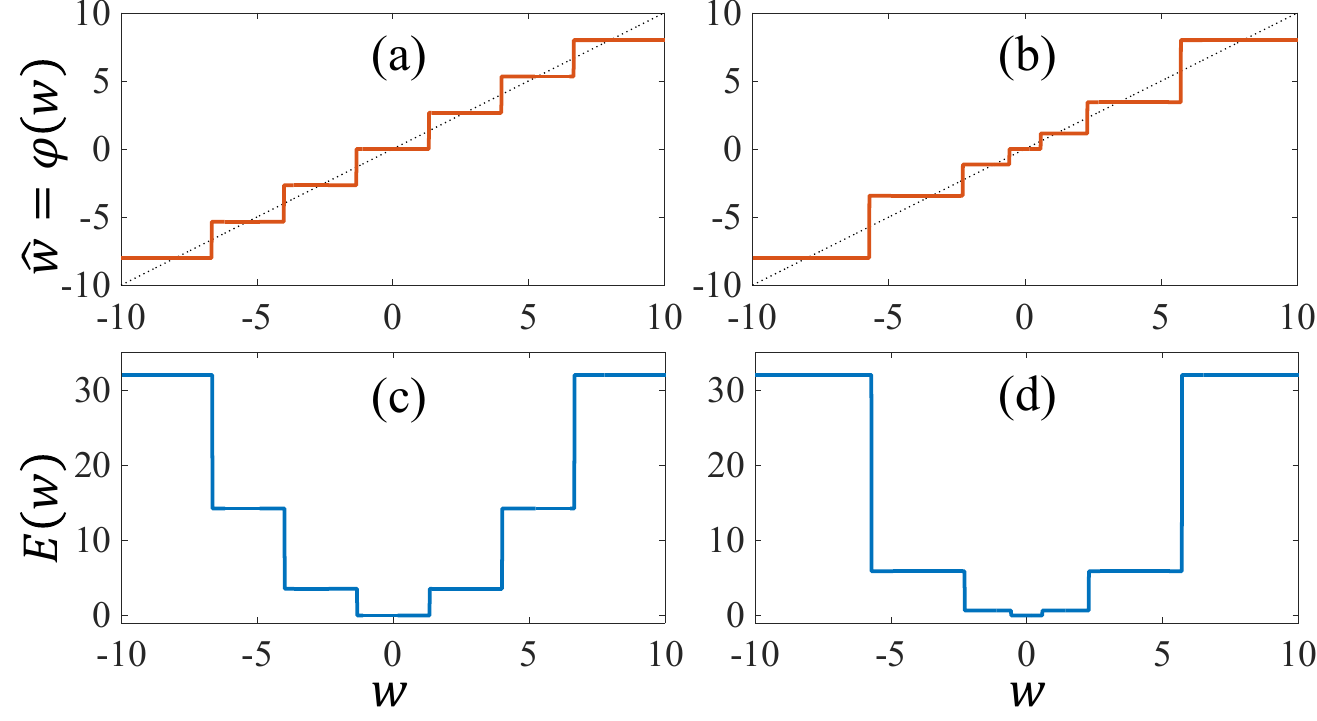}
    \caption{(a) and (b): Quanzied values $\widehat{w}=\varphi(w)$ as a function of continuous parameter $w$ at $\omega=8$ and $n_p=6$ for (a) uniform and (b) non-uniform cases.
    The diagonal lines represent the identity map.
    (c) and (d): Loss function at $N=M=1$ for $y=0$ and $x=1$ with $\lambda=0$ for (c) uniform and
    (d) non-uniform quantization
    corresponding to (a) and (b), respectively.}
    \label{fig:quantization}
\end{figure}
\subsection{Quantization}

Quantization is a procedure that converts real values $w\in\mathbb{R}$ into their closest discrete counterparts $\widehat{w}$ within a predefined set of candidate values $\Omega$. We represent the quantization as $\widehat{w}=\varphi(w)$, using a quantization function $\varphi:\mathbb{R}\to\Omega$ that outputs the value in $\Omega$ closest to $w$.
This quantization process is determined by clipping and partitioning \cite{wu2020integer}. In the clipping process, continuous values whose absolute magnitude exceeds $\omega$ are capped at $\pm\omega$, while their respective signs are preserved. 
Further, the range $[-\omega,\omega]$ is divided into $n_p$ subintervals following the partitioning procedure, and possible discrete values are located at the edges of these subintervals, hence $|\Omega|=n_p+1$. Here, we define the number of bits, $b := \log_2(n_p + 2)$, capable of assuming real values. We investigate two types of quantization:
uniform and non-uniform quantizations,
which are illustrated in Fig.\ref{fig:quantization} (a) and (b).

\subsubsection{Uniform Quantization \cite{wu2020integer}}

In uniform quantization, the 
subintervals have equal width ${2\omega} / {n_p}$. 

\subsubsection{Non-Uniform Quantization}
Non-uniform quantization is introduced to finely partition the significant value regions \cite{han2015deep,zhu2016trained}. We consider a non-uniform quantization based on a uniform partition in the log-domain, which is a slightly updated version of \cite{miyashita2016convolutional,zhou2016incremental}. Here, the subinterval width closest to zero is set to 
$\Delta_0$ given by,
\begin{align}
  \Delta_0=\frac{\omega}{2^{({n_p+k})/{2}}-k}
\end{align}
where $k=2$ for an even $n_p$ and $k=3$ for an odd $n_p$.
The widths of other subintervals are larger than $\Delta_0$ and are twice as wide as their nearest neighbor subinterval width toward zero\footnote{We define this partition width using $n_p$ and $\omega$ to compare it with uniform quantization. For various designs of non-uniform quantization, see the comprehensive review by \cite{baskin2021uniq}.}.

\subsection{Learning Problem with Linear Model}
We formulate the problem setting.
Let $M$ be a sample size and $N$ be a dimension.
Suppose that we have $M$ pairs of an predictor and output $\mathcal{D}:=\{(\bm{x}_{\mu}, y_{\mu})\}_{\mu=1}^M \subset \mathbb{R}^N \times \mathbb{R}$, 
which is independently generated from the linear model $y_{\mu} = \bm{x}_{\mu}^\top{\bm{w}^0} + \varepsilon_{\mu}^0$ with the true parameter $\bm{w}^0\in\mathbb{R}^{N}$ and the independent noise $\varepsilon_\mu^0\sim{\mathcal N}(0,\sigma^2)$. 
The components of the true parameter and predictors are independently drawn from ${\mathcal N}(0,\rho)$ with $\rho > 0$ and ${\mathcal N}(0,1\slash N)$, respectively \footnote{These settings are not restrictive as it seems. It has been shown that asymptotic behaviors of Gaussian universality regression problems with a large class of features can be computed to leading order under a simpler model with Gaussian features \cite{loureiro2021learning,goldt2022gaussian,montanari2022universality,gerace2023gaussian}.}.
We also introduce a notation $\bm{y}=(y_1,...,y_M)^\top \in\mathbb{R}^M$ and $\bm{X}=(\bm{x}_1,...,\bm{x}_M)^\top \in \mathbb{R}^{M \times N}$.

We formulate the estimation procedure for parameters with quantization.
We define a quantization map $\bm{\varphi}: \mathbb{R}^N \to \Omega^N$ for a vector $\bm{w} = (w_1,...,w_N) \in \mathbb{R}^N$ as $\bm{\varphi}(\bm{w})=(\varphi(w_1) , ... , \varphi(w_N))^\top$.
Then, we consider the empirical risk of $\bm{w} \in \mathbb{R}^N$ with the $\ell_2$ regularization with a coefficient $\lambda>0$:

\begin{align}
E(\bm{w};{\mathcal D}) &:=  \frac{1}{2}
\|\bm{y}-\bm{X}\bm{\varphi}(\bm{w})\|_2^2
+ \frac{\lambda}{2} \|\bm{\varphi}(\bm{w})\|_2^2.
\end{align}

The loss function is nonconvex due to the quantization, as shown in Fig.\ref{fig:quantization}.
The trained continuous parameter
\begin{align}
\bm{w}^*({\mathcal D}) &:= \argmin_{\bm{w} \in \mathbb{R}^N} E(\bm{w}; \mathcal D),\label{eq:def_estimate}
\end{align}
constitute continuous sets 
that cover subinterval at the bottom of loss (Fig.\ref{fig:quantization} (c) and (d)).
Here, we denote quantized values of \eqref{eq:def_estimate} as
$\widehat{\bm{w}}^*({\mathcal D}):=\bm{\varphi}(\bm{w}^*({\mathcal D}))$.

Our interest is a \textit{generalization error}, which measures a prediction performance of the trained parameter $\bm{w}^*({\mathcal D})$.
With a newly generated pair $(y_{\mathrm{new}},\bm{x}_{\mathrm{new}})$ from the identical distribution of the observations $\mathcal{D}$, we define an expectation of the generalization error as

\begin{align}
    \overline{E}_g :=\frac{1}{2} \mathbb{E}_{{\mathcal D}}[\mathbb{E}_{(y_{\mathrm{new}},\bm{x}_{\mathrm{new}})}[(y_{\mathrm{new}}-\bm{x}_{\mathrm{new}}^\top\widehat{\bm{w}}^*({\mathcal D}))^2]].
    \label{eq:def_generalization}
\end{align}
Simple calculations yield the expression 
\begin{align}
    \overline{E}_g=\frac{1}{2}\left(Q_\varphi-2m_\varphi+\rho+\sigma^2\right) \label{eq:gen_err}
\end{align}
with $Q_\varphi:=\frac{1}{N}\mathbb{E}_{{\mathcal D}}[\|\widehat{\bm{w}}^*({\mathcal D})\|_2^2]$ and $m_\varphi:=\frac{1}{N}\mathbb{E}_{{\mathcal D}}[{\bm{w}^0}^\top\widehat{\bm{w}}^*({\mathcal D})]$.


\section{Replica Analysis}
We employ \textit{replica method} \cite{Mezard1987,SpinGlassFarBeyond} to compute the terms $Q_\varphi$ and $m_\varphi$  in \eqref{eq:gen_err}.
In preparation,
we introduce the posterior distribution with a parameter $\beta > 0$ as
\begin{align}
    p_{\beta}(\bm{w}|\mathcal{D})= \exp(-\beta E(\bm{w};{\mathcal D}))/Z_{\beta}(\mathcal{D}),
    \label{eq:posterior}
\end{align}
where $Z_\beta({\mathcal D})=\int d\bm{w}\exp(-\beta E(\bm{w}))$
is the normalization constant.
At $\beta\to\infty$, the posterior distribution 
converges to the uniform distribution over the solution \eqref{eq:def_estimate}.
Denoting the posterior mean
at $\beta\to\infty$ as $\langle\bm{w}\rangle_{\mathcal D}$,
its quantized value $\bm{\varphi}(\langle\bm{w}\rangle_{\mathcal D})$ is equivalent to $\widehat{\bm{w}}^*({\mathcal D})$, when 
the quantized value 
of \eqref{eq:def_estimate} is unique.
Further, $\langle \varphi(\bm{w})\rangle_{\mathcal D}=\varphi(\langle\bm{w}\rangle_{\mathcal D})$ holds, hence we obtain 
$Q_\varphi
=\frac{1}{N}\mathbb{E}_{{\mathcal D}}\left[\langle \|\varphi(\bm{w})\|_2^2
\rangle_{\mathcal D}\right]$ and 
$m_\varphi
=\frac{1}{N}\mathbb{E}_{{\mathcal D}}[{\bm{w}^0}^\top
\!\langle\bm{\varphi}(\bm{w})\rangle_{\mathcal D}]$. 
Since the posterior mean has  $Z_\beta({\mathcal D})$
in its denominator, which
requires summations over exponential number of terms, we need the following procedure.
As we set 
\begin{align}
    Q_\varphi(n)=
    \lim_{\beta\to\infty}\!\frac{1}{N}\mathbb{E}_{{\mathcal D}}\!\left[Z_\beta^n({\mathcal D})\frac{\int d\bm{w}
    \|\varphi(\bm{w})\|_2^2\exp(-\beta E(\bm{w}))}{Z_\beta({\mathcal D})}\right],
\end{align}
it is obvious that 
$Q_\varphi=\lim_{n\to 0}Q_\varphi(n)$ holds.
Assuming that $n$
is an integer larger than 1,
the denominator is canceled,
and the term $Z_\beta^{n-1}({\mathcal D})\exp(-\beta E(\bm{w}))$
can be expressed with 
$n$-replicated systems with $\bm{w}^1,\cdots,\bm{w}^n$.
At $M\to\infty$
and $N\to\infty$ with keeping $\alpha=M\slash N\sim O(1)$,
the expectation 
$\mathbb{E}_{\mathcal D}[\cdot]$
is implemented by the saddle point evaluation.
The resulting expression for the integer $n$ 
is analytically continued to $n\in\mathbb{R}$ 
to take the limit $n\to 0$
under the replica symmetric (RS) ansatz, 
where we assume that the
dominant saddle
point 
is invariant under any permutation of replica indices $a=1,2,\cdots,n$.
Finally, $Q_\varphi$ and $m_\varphi$
are given by $Q$ and $m$ that satisfy saddle point equations
\begin{align}
Q = \int Dz ({\varphi^*(hz,\widehat{\Theta})})^2,~~
m =\hat{m}\rho\int Dz 
\partial \varphi^*(hz,\widehat{\Theta}),
\label{eq:saddle_point_m}
\end{align}
where $Dz=\frac{dz}{\sqrt{2\pi}}
e^{-\frac{1}{2}z^2}$, 
$h=\sqrt{\widehat{m}^2\rho+\widehat{\chi}}$ and
$\widehat{\Theta}=\widehat{Q}+\lambda$
satisfy
\begin{align}
\hat{Q}&=\hat{m}= \frac{\alpha}{1+\chi},~~
\hat{\chi} = \frac{2\alpha\overline{\mathcal E}_g}{(1+\chi)^2},~~
\chi =\!\! \int \!\!Dz \partial \varphi^*(hz,\widehat{\Theta})
.\label{saddle_point_chi}
\end{align}
Here, $\overline{\mathcal E}_g = \frac{1}{2}(Q-2m+\rho + \sigma^2)$ coincides with $\overline{E}_g$ at the saddle point.
The function $\varphi^* (hz,\widehat{\Theta})$ is the solution of the problem
\begin{align}
\varphi^*(hz,\widehat{\Theta}) = \argmin_{d\in \Omega} \left\{\frac{1}{2}\widehat{\Theta}d^2 - hzd\right\},
\label{eq:effective_single_body}
\end{align}
which corresponds $\varphi(hz\slash\widehat{\Theta})$, namely the 
the solution of the quantization-restricted ridge optimization problem under RS assumption\footnote{
This correspondence does not mean that the obtained quantized posterior mean is equivalent to the quantized solution for ridge regression. The saddle points depend on the quantization function, hence the values of $h$ and $\widehat{\Theta}$ for quantized regression differ from those in usual ridge regression.}\cite{kabashima2009typical}, and $\partial\varphi^*(hz,\widehat{\Theta})=\frac{\partial}{\partial (hz)}\varphi^*(hz,\widehat{\Theta})$.
The random variable $z$ effectively represents 
the randomness induced by data ${\mathcal D}$,
and the solution of \eqref{eq:effective_single_body} is statistically equivalent to the 
solution of the original problem \eqref{eq:def_estimate} \cite{Bayati-Montanari2011}.

\begin{figure}
    \centering
    \includegraphics[width=150mm]
    {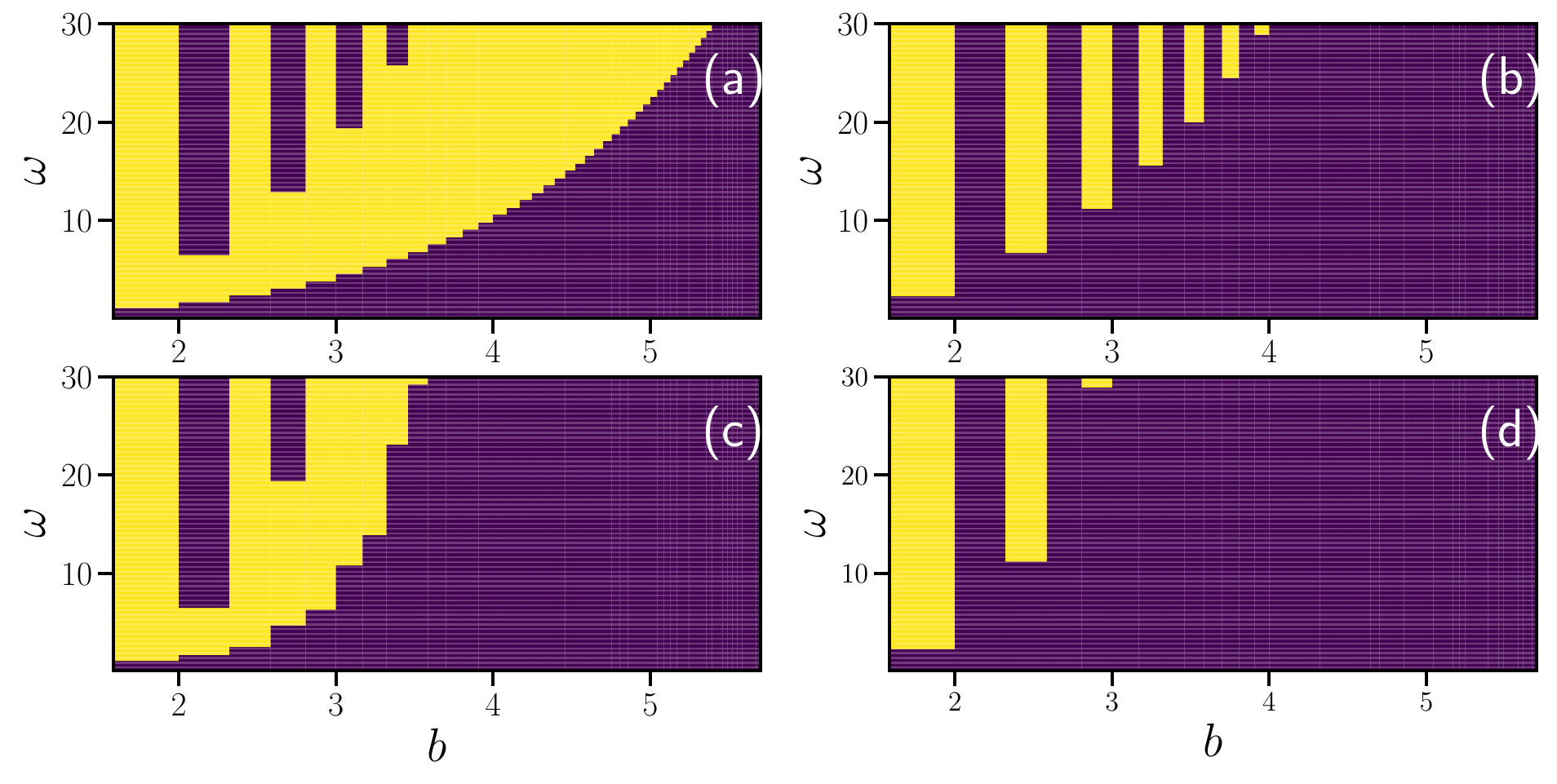}
    \caption{Phase diagrams
    on $b-\omega$ plane
    for quantized regression at $(\sigma, \alpha)=(0.01,1.5)$.
    Shaded and blighted regions are RS and RSB phase, respectively.
    (a) and (b) are under uniform quantization at $\lambda = 0.0$ and $\lambda = 1.0$, respectively. (c) and (d) are under non-uniform quantization at $\lambda = 0.0$ and $\lambda = 1.0$, respectively.}
    \label{fig:phase_diagram}
\end{figure}
\section{Result}

\subsection{Phase Diagram}


We study the RS ansatz for replica analysis.
The RS solution loses local stability against symmetry breaking perturbations
when the following condition is not satisfied 
\cite{Almeida1978}:
\begin{align}
\frac{\alpha}{(1+\chi)^2} \int Dz
\left(\partial \varphi^*(hz,\widehat{\Theta})\right)^2
< 1. \label{def:stability_condition}
\end{align}
The phases satisfying \eqref{def:stability_condition}
and 
that not satisfying it
are termed as RS and replica symmetry breaking (RSB) phases, respectively.
In the RSB phase, 
algorithmic instabilities arise due to exponential number of local minima \cite{auer1995exponentially,Mezard1987}, necessitating quantization hyperparameter setting to avoid the RSB phase.




Figure \ref{fig:phase_diagram} shows the phase diagrams 
for uniform quantization ((a), (b)) and non-uniform quantization ((c), (d)). 
We obtain several implications as follows.

(I): Irrespective of the quantization method, large bits $b$ and small $\omega$ result in RS phase.
This aligns with the fact that a large $b$ makes quantized values closer to continuous values. Additionally, a small 
$\omega$ induces shrinkage in estimates, which tends to be the RS phase as the effective model complexity is reduced  \cite{zou2007degree,sakata2023prediction}.

(II): The stronger regularization with large $\lambda$ makes the RS phase larger. 
This is consistent with the general trend that regularization improves the stability of the model by reducing the variance of weights.
\begin{figure}
    \centering
    \includegraphics[width=150mm]{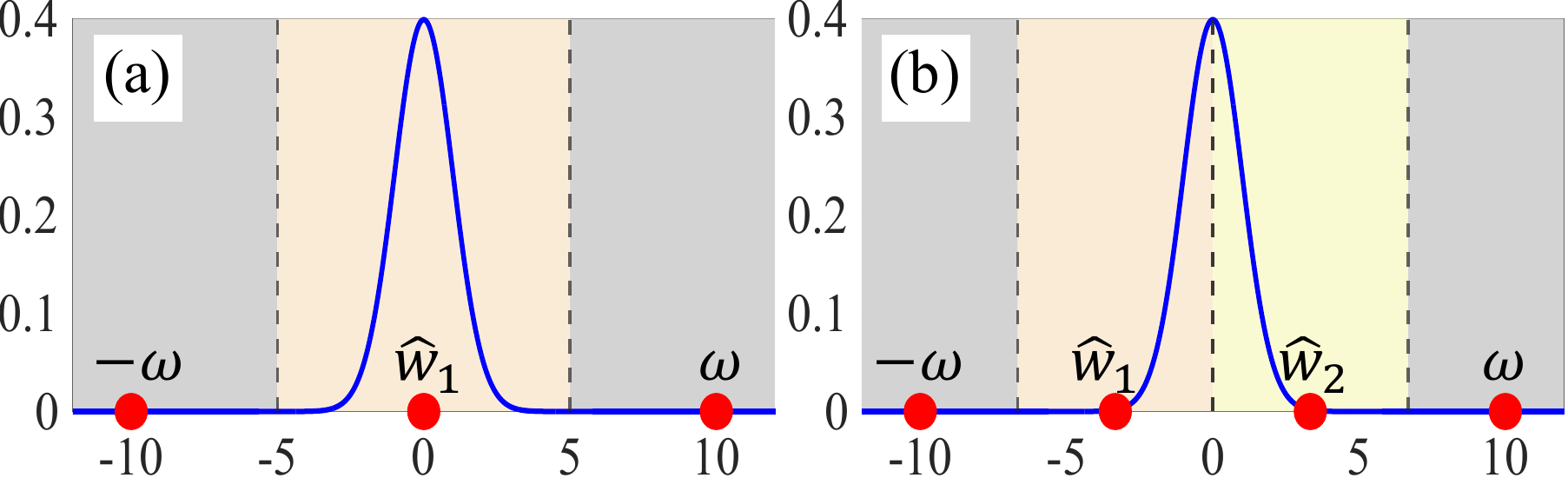}
    \caption{Comparison between the distribution of continuous value $hz\slash\widehat{\Theta}$ (solid line) and quantized values in $\Omega$ with $\omega=10$ (dots) for (a) $n_p=2$ and (b) $n_p=3$ under uniform quantization. Here, we set $h\slash\widehat{\Theta}=1$ for simplicity, and areas corresponding to the discrete values are shaded separately.}
    \label{fig:single_body_problem}
\end{figure}
(III): In regions with small bits $b$ and large range $\omega$, 
the RS and RSB phases alternate periodically along the $b$ direction, resulting in a striped phase diagram.
The striped pattern is caused by the dependence of $\Omega$ on $n_p$ (or $b$), leading to RS for even $n_p$ and RSB for odd $n_p$.
As illustrated in Fig. \ref{fig:single_body_problem}, for even $n_p$, $\Omega$ includes zero ($\widehat{w}_1$ of (a)), while for odd $n_p$, it does not. For large $\omega$ with small $n_p$, the intervals between quantized values can be substantial compared to $h\slash\widehat{\Theta}$, which represents the variance of the continuous value to be quantized in \eqref{eq:effective_single_body}.
Hence, for even $n_p$, continuous values are highly likely to be quantized to zero, 
while for odd $n_p$, they tend to be quantized to the non-zero value closest to zero ($\widehat{w}_1$ or $\widehat{w}_2$ in Fig.\ref{fig:single_body_problem} (b) where $|\widehat{w}_1|=|\widehat{w}_2|$).
The closest values can be substantial relative to $h\slash\widehat{\Theta}$ at large $\omega$ and small $n_p$, causing the quantization at odd $n_p$ to extend the continuous value. This parameter extension increases effective degrees of freedom \cite{zou2007degree}, resulting in the instability of the RS solution.

(IV): The non-uniform quantization exhibits a larger RS phase compared to uniform one, indicating a broader range for quantization settings.
In non-uniform quantization, the subintervals near the origin are small, effectively suppressing the extension caused by quantization for odd $n_p$, leading to a wider RS phase.

\begin{figure}[t]
    \begin{centering}
  \includegraphics[width = 150mm]{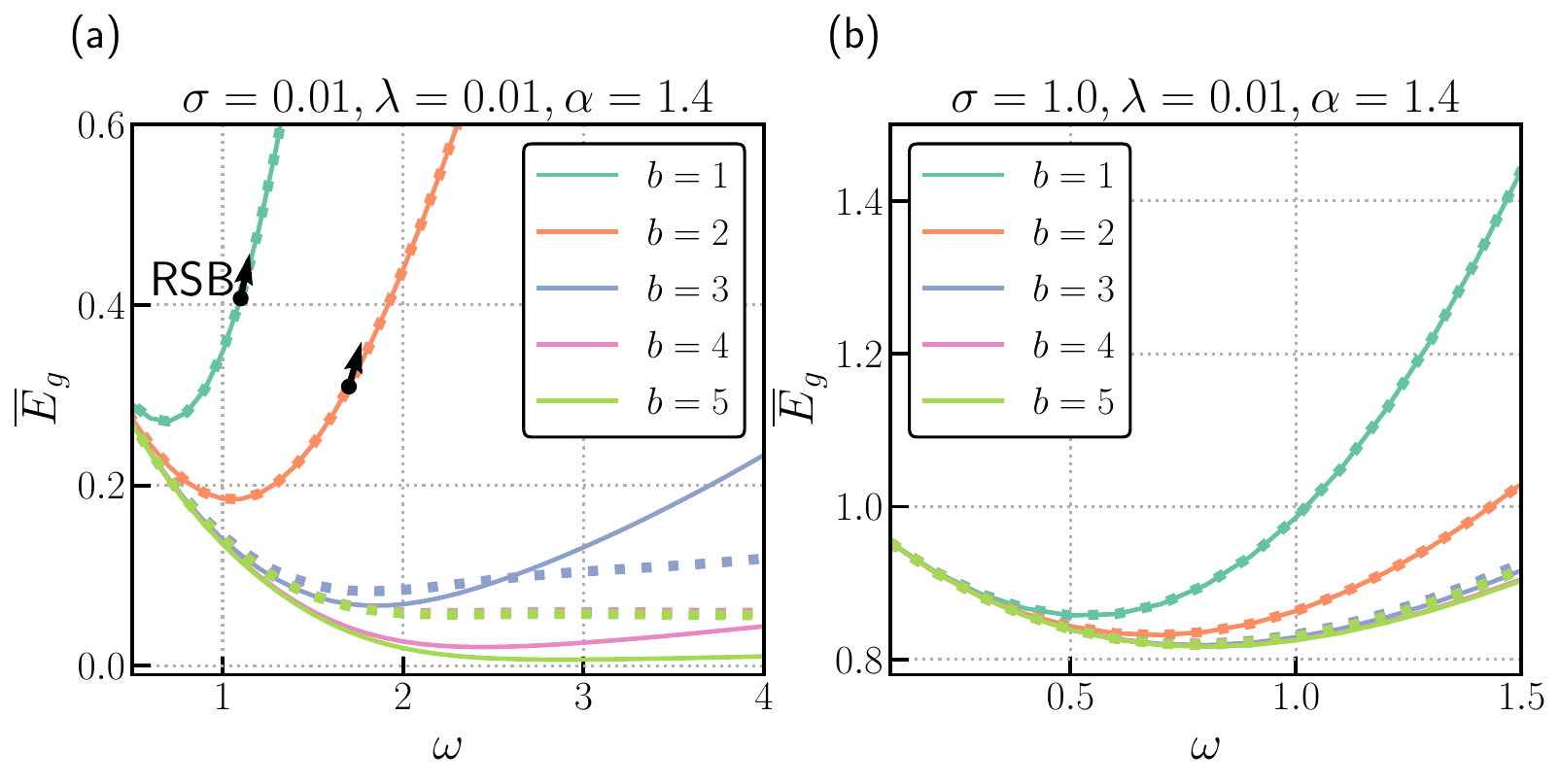}
      \end{centering}
      \caption{
     Expected generalization error under RS assumption as a function of $\omega$ for (a) $(\sigma, \lambda, \alpha)=(0.01, 0.01, 1.4)$ and (b) $(\sigma, \lambda, \alpha)=(1.0, 0.01, 1.4)$.
Solid lines and dashesd lines represent the result of uniform and non-uniform quantization, respectively. RSB phase is indicated by black arrows. 
      }
      \label{fig_2_a_b}
\end{figure}

\subsection{Optimal Parameters in RS Phase}



We now investigate the generalization error $\overline{E}_g$ in RS phase to consider optimal quantization hyperparameters.
Figure \ref{fig_2_a_b} shows the generalization errors under RS ansatz with different values of $b$ and $\omega$ for (a) low and (b) high noise cases. 
Insights derived from the behavior of generalization errors are as follows.
First, in most cases, the generalization error is a convex function of $\omega$ and has a unique minimum. 
This characteristics implies that there exists the optimal choice for the range $\omega$, which should be appropriately selected.
Especially when the number of bits is small, $\overline{E}_g$
exhibits a stronger dependence on 
$\omega$ compared to cases with a larger number of bits, resulting in a sharper minimum value of $\overline{E}_g$. Consequently, the appropriate selection of 
$\omega$ becomes more crucial for scenarios with a small number of bits.
Second, as the noise $\sigma$ increases with the large bit $b$ increases, the generalization error of non-uniform and uniform quantization disappears. 



\subsection{Effect on Double Descent}
\begin{figure}[t]
    \begin{centering}
  \includegraphics[width = 150mm]{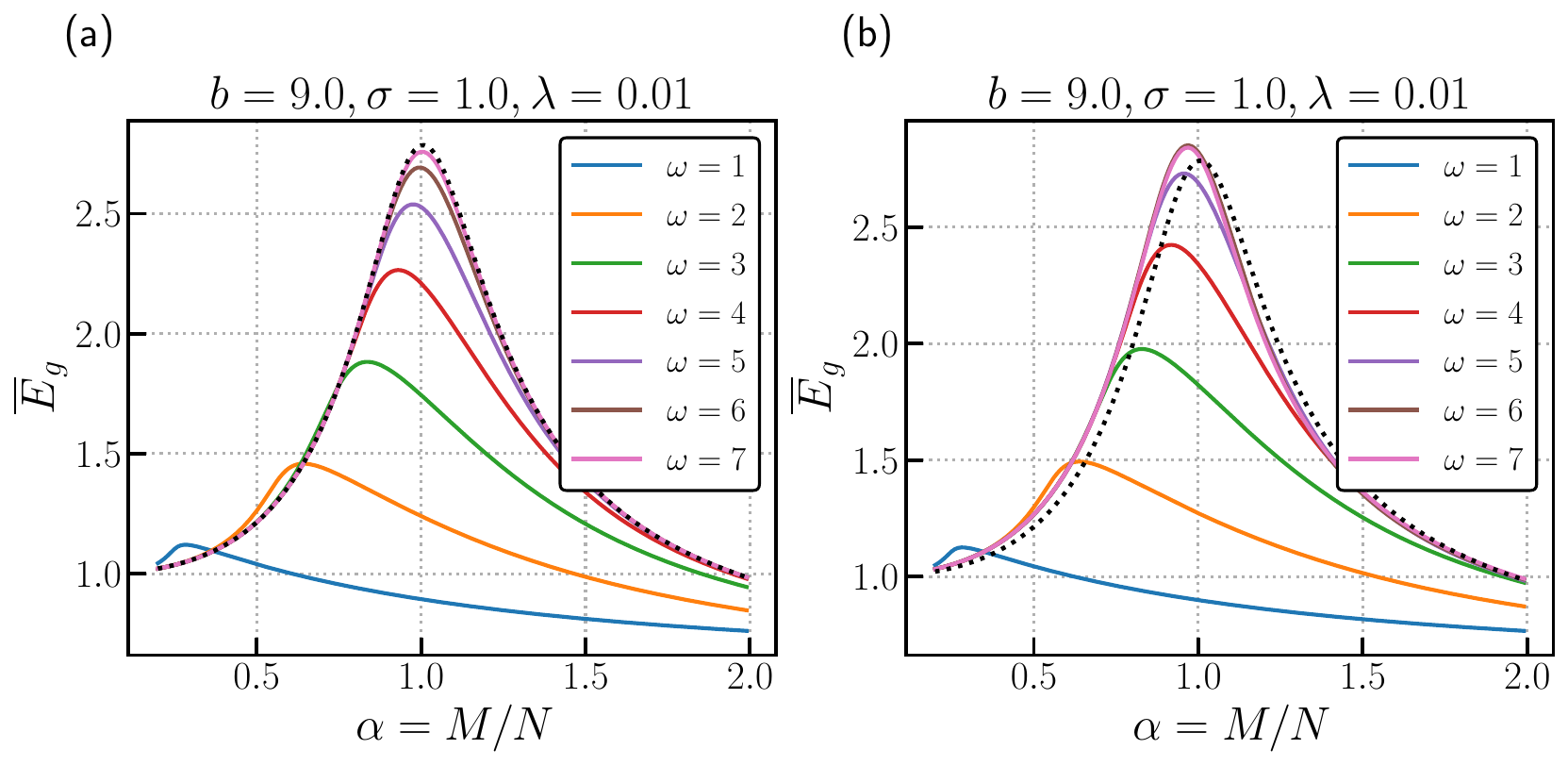}
      \end{centering}
      \vskip -0.2in
      \caption{
      Expected generalization error under RS assumption as a function of $\alpha = N/M$ for (a) uniform quantization and (b) non-uniform quantization.
The dashed lines represent the result of ridge regression.}
      \label{fig_3_a_b}
\end{figure}

We investigate the effect of quantization on the double-descent phenomenon of the generalization error $\overline{E}_g$. 
The double-descent refers to the phenomenon in which the generalization error increases once and then decreases when the number of parameters is extremely large\cite{belkin2019reconciling, loog2020brief}. 
This is an important notion in modern data analysis, since it suggests that an excess number of parameters enters a regime where the low performance from overfitting can be avoided \cite{belkin2019reconciling,hastie2022surprises}.
Ridge regression with continuous parameter exhibits the double descent phenomena, where generalization error has a peak at $\alpha=M\slash N=1$ under sufficiently small $\lambda$ and sufficiently large $\sigma$ \cite{hastie2022surprises,Krogh1992}.

Fig.\ref{fig_3_a_b} shows 
the generalization errors under quantization as a function of $\alpha$
for (a) uniform and (b) non-uniform cases.
The quantization shifts the peak toward smaller $\alpha$ compared with the usual ridge regression, which is indicated by dashed line. 
This means that more parameters are needed to reach the regime of overparameterization under the quantization. 
This implies that it is slightly difficult to benefit from the overparameterization, since the quantization reducing an effective number of parameters. 
This tendency is marginally more pronounced for non-uniform quantization.
However, this discrepancy disappears with the larger $\omega$.

\section{AMP Algorithm and Experiments}
\begin{figure}[t]
    \begin{centering}
  \includegraphics[width = 150mm]{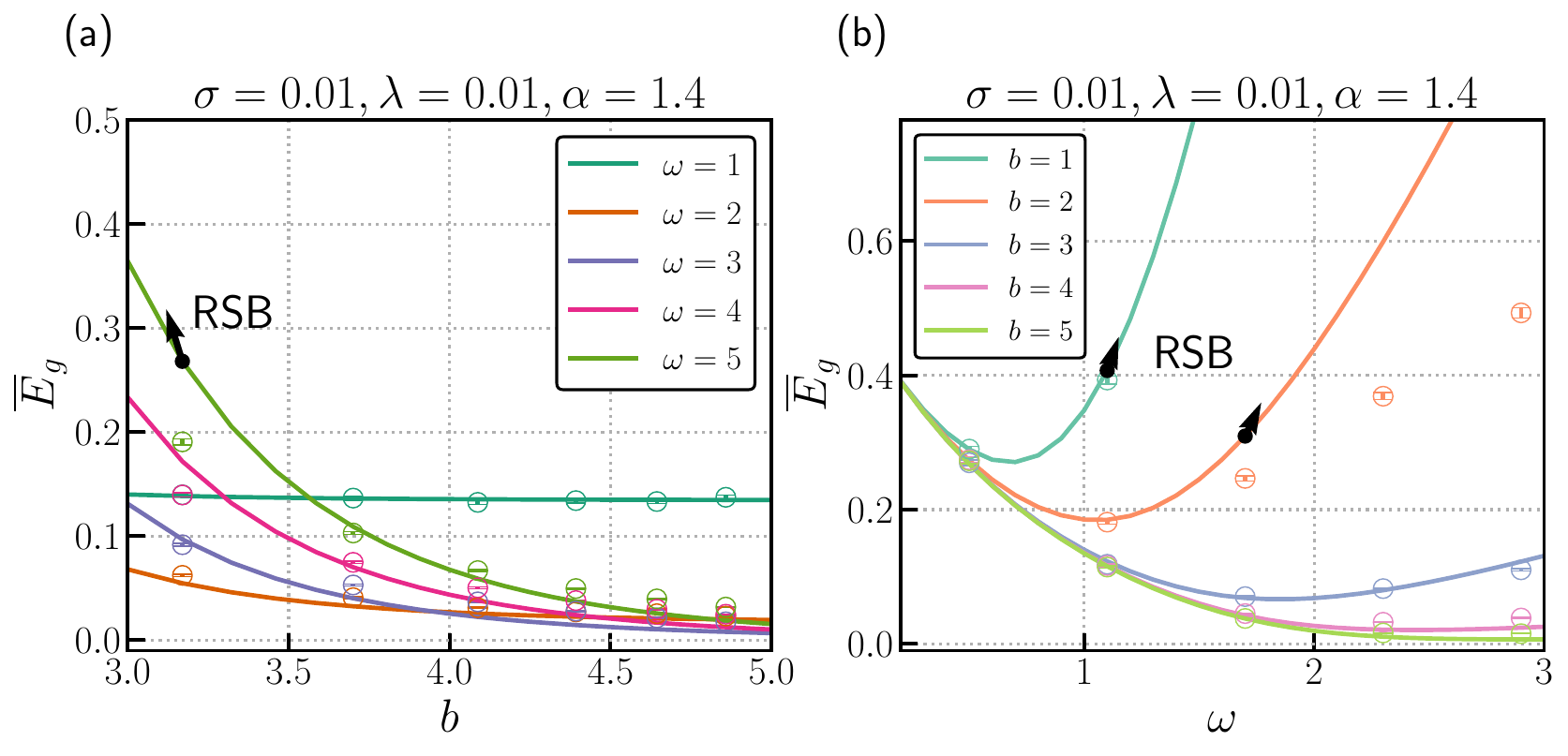}
      \end{centering}
            \vskip -0.2in
      \caption{Comparison of results by replica analysis (solid line) and AMP ($\circ$) in terms of generalization error dependence on (a) $b$ and (b) $\omega$.
The error bars denote standard errors assessed by 100 independent runs of AMP, where the each component of initial conditions $\overline{\bm{m}}^0$ is generated by ${\mathcal N}(0,1)$.
RSB phase is indicated by black arrows. }
      \label{fig_4_a_b}
\end{figure}
As a numerical method that can potentially achieve the theoretical results under the replica symmetry (RS) ansatz, we employ the approximate message passing (AMP) algorithm
\cite{Donoho2009,Rangan2011}.
AMP for quantized regression includes the procedures outlined in Algorithm \ref{alg:AMP}, which utilize a quantization function \eqref{eq:effective_single_body} and its derivative. 
As in replica analysis,
the quantized estimate by AMP in step $t$, $\bar{\bm{m}}^t$,
corresponds to the quantized solution of the ridge regression \cite{sakata2021perfect}.
In the quantized regression,
the following relationship between AMP and replica analysis holds,
as with continuous regression \cite{Zdeborova2016,Krzakala2012PRX}.
First, the typical trajectory of AMP can be described by the state evolution equation,
which corresponds to the saddle point equations derived by replica analysis under the RS ansatz.
Second, the microscopic convergence condition of AMP corresponds to the stability condition of the RS solution \eqref{def:stability_condition}.
This property indicates that the
theoretical result for quantized regression can be achieved by AMP, in principle.

Fig. \ref{fig_4_a_b} shows $b$ and $\omega$-dependence of  $\overline{E}_g$ at $N=2500$, computed with the form \eqref{eq:def_generalization} using AMP's estimate $\bar{\bm{m}}^t$ at sufficiently large $t$ as $\widehat{\bm{w}}^*({\mathcal D})$.
The observed numerical results are in good agreement with the theory in the RS phase, and indicate the validity of the AMP for the quantized regression problems.

\begin{algorithm}[ht]
\begin{algorithmic}
\STATE{\textbf{Initialize} $\{(V_\mu^0, \theta_\mu^0)\}_{\mu=1}^M$, $\{(\Sigma_i^0, R_i^0)\}_{i=1}^N, \bar{\bm{m}}^0, {\bm{v}}^0$.}
\FOR{$t=1,...,T$}
\STATE{$V_{\mu}^t \gets (\bm{v}^{t-1})^\top\bm{x}_{\mu}$}
\STATE{$(\Sigma_i^t)^{-1}\gets\sum_{\mu=1}^M {X_{\mu i}^2}\slash(V_{\mu}^t+1)$ }
\STATE{$\theta_{\mu}^t \gets
(\bar{\bm{m}}^{t-1})^\top\bm{x}_\mu
-{V_{\mu}^t(y_{\mu}-\theta_{\mu}^{t-1})}\slash({V_{\mu}^{t-1}+1}),$}
\STATE{$R_i^t\gets\bar{m}_i^{t-1}+\Sigma_i^t\sum_{\mu=1}^M X_{\mu i}(y_{\mu}-\theta_{\mu}^t)/(V_{\mu}^{t-1}+1)$}
\STATE{$\bar{m}_i^{t}\gets\varphi^*(R_i^t\slash\Sigma_i^t, \lambda+\!1\slash\Sigma_i^t)$}
\STATE{$v_i^t\gets\partial \varphi^* (R_i^t\slash\Sigma_i^t, \lambda+
\!1\slash\Sigma_i^t)$}
\ENDFOR
\end{algorithmic}
\caption{AMP for quantized regression \\ ~~
$\bar{\bm{m}}^t=\{\bar{m}_i^{t}\}_{i=1}^N$, $\bm{v}^t=\{v_i^t\}_{i=1}^N$, $\bm{x}_\mu = \{ X_{\mu  i}\}_{i=1}^N$.}
\label{alg:AMP}
\end{algorithm}




\section{Discussion and Conclusion}

In this study, we provided the theoretical results of the effect of hyperparameters of quantization.
We investigated two types of quantization: uniform and non-uniform.
First, we derived the RS-RSB phase diagram that describes the validity of our theoretical results. 
We showed that non-uniform quantization and regularization can reduce the extent of the RSB phase.
Second, we demonstrated that an optimal quantization range exists, which minimizes the generalization error.
Lastly, we found that more parameters are required to diminish the generalization error in the overparametrization regime when applying quantization.
Our numerical experiments by AMP are in good agreement with theoretical results.

It is reported that AMP fail to achieve the theoretical limit for nonconvex problems \cite{sakata2021perfect}.
Investigating the existence of this divergence in quantized regression is future work.
Additionally, extending our results to develop a practical method for selecting hyperparameters is also the subject of future work.




\section*{Acknowledgment}

SK is supported by JSPS KAKENHI (23KJ0723) and JST CREST (JPMJCR21D2).
AS is supported by JSPS KAKENHI (22H05117) and JST PRESTO (JPMJPR23J4).
MI is supported by JSPS KAKENHI (21K11780), JST CREST (JPMJCR21D2), and JST FOREST (JPMJFR216I).

\bibliographystyle{ieeetr}
\bibliography{preprint}

\newcommand{\etalchar}[1]{$^{#1}$}
\providecommand{\noopsort}[1]{}\providecommand{\singleletter}[1]{#1}%
\begin{thebibliography}{KMS{\etalchar{+}}12b}

\bibitem[AHW95]{auer1995exponentially}
Peter Auer, Mark Herbster, and Manfred~KK Warmuth.
\newblock Exponentially many local minima for single neurons.
\newblock {\em Advances in neural information processing systems}, 8, 1995.

\bibitem[BGL{\etalchar{+}}16]{baldassi2016}
Carlo Baldassi, Federica Gerace, Carlo Lucibello, Luca Saglietti, and Riccardo
  Zecchina.
\newblock Learning may need only a few bits of synaptic precision.
\newblock {\em Physical Review E}, 93:052313, 2016.

\bibitem[BHHS18]{banner2018scalable}
Ron Banner, Itay Hubara, Elad Hoffer, and Daniel Soudry.
\newblock Scalable methods for 8-bit training of neural networks.
\newblock {\em Advances in neural information processing systems}, 31, 2018.

\bibitem[BHMM19]{belkin2019reconciling}
Mikhail Belkin, Daniel Hsu, Siyuan Ma, and Soumik Mandal.
\newblock Reconciling modern machine-learning practice and the classical
  bias--variance trade-off.
\newblock {\em Proceedings of the National Academy of Sciences},
  116(32):15849--15854, 2019.

\bibitem[BLS{\etalchar{+}}21]{baskin2021uniq}
Chaim Baskin, Natan Liss, Eli Schwartz, Evgenii Zheltonozhskii, Raja Giryes,
  Alex~M Bronstein, and Avi Mendelson.
\newblock Uniq: Uniform noise injection for non-uniform quantization of neural
  networks.
\newblock {\em ACM Transactions on Computer Systems (TOCS)}, 37(1-4):1--15,
  2021.

\bibitem[BM11]{Bayati-Montanari2011}
Mohsen Bayati and Andrea Montanari.
\newblock The dynamics of message passing on dense graphs, with applications to
  compressed sensing.
\newblock {\em IEEE Transaction on information theory}, 57(2):764--785, 2011.

\bibitem[BMR{\etalchar{+}}20]{brown2020language}
Tom Brown, Benjamin Mann, Nick Ryder, Melanie Subbiah, Jared~D Kaplan, Prafulla
  Dhariwal, Arvind Neelakantan, Pranav Shyam, Girish Sastry, Amanda Askell,
  et~al.
\newblock Language models are few-shot learners.
\newblock {\em Advances in neural information processing systems},
  33:1877--1901, 2020.

\bibitem[CBD14]{courbariaux2014training}
Matthieu Courbariaux, Yoshua Bengio, and Jean-Pierre David.
\newblock Training deep neural networks with low precision multiplications.
\newblock {\em arXiv preprint arXiv:1412.7024}, 2014.

\bibitem[CBUS{\etalchar{+}}20]{chmiel2020neural}
Brian Chmiel, Liad Ben-Uri, Moran Shkolnik, Elad Hoffer, Ron Banner, and Daniel
  Soudry.
\newblock Neural gradients are near-lognormal: improved quantized and sparse
  training.
\newblock In {\em International Conference on Learning Representations}, 2020.

\bibitem[CEKL16]{choi2016towards}
Yoojin Choi, Mostafa El-Khamy, and Jungwon Lee.
\newblock Towards the limit of network quantization.
\newblock In {\em International Conference on Learning Representations}, 2016.

\bibitem[CMP{\etalchar{+}}23]{SpinGlassFarBeyond}
Patrick Charbonneau, Enzo Marinari, Giorgio Parisi, Federico Ricci-tersenghi,
  Gabriele Sicuro, Francesco Zamponi, and Marc Mezard.
\newblock {\em Spin Glass Theory and Far Beyond: Replica Symmetry Breaking
  after 40 Years}.
\newblock World Scientific, 2023.

\bibitem[DAT78]{Almeida1978}
JRL De~Almeida and David~J Thouless.
\newblock Stability of the sherrington-kirkpatrick solution of a spin glass
  model.
\newblock {\em Journal of Physics A: Mathematical and General}, 11(5):983,
  1978.

\bibitem[DLXS19]{ding2019universal}
Yukun Ding, Jinglan Liu, Jinjun Xiong, and Yiyu Shi.
\newblock On the universal approximability and complexity bounds of quantized
  relu neural networks.
\newblock In {\em International Conference on Learning Representations}.
  International Conference on Learning Representations, ICLR, 2019.

\bibitem[DMM09]{Donoho2009}
David~L Donoho, Arian Maleki, and Andrea Montanari.
\newblock Message-passing algorithms for compressed sensing.
\newblock {\em Proceedings of the National Academy of Sciences},
  106(45):18914--18919, 2009.

\bibitem[GBGR23]{gonon2023approximation}
Antoine Gonon, Nicolas Brisebarre, R{\'e}mi Gribonval, and Elisa Riccietti.
\newblock Approximation speed of quantized vs. unquantized relu neural networks
  and beyond.
\newblock {\em IEEE Transactions on Information Theory}, 2023.

\bibitem[GKD{\etalchar{+}}22]{gholami2022survey}
Amir Gholami, Sehoon Kim, Zhen Dong, Zhewei Yao, Michael~W Mahoney, and Kurt
  Keutzer.
\newblock A survey of quantization methods for efficient neural network
  inference.
\newblock In {\em Low-Power Computer Vision}, pages 291--326. Chapman and
  Hall/CRC, 2022.

\bibitem[GKL{\etalchar{+}}23]{gerace2023gaussian}
Federica Gerace, Florent Krzakala, Bruno Loureiro, Ludovic Stephan, and Lenka
  Zdeborov{\'a}.
\newblock Gaussian universality of perceptrons with random labels.
\newblock 2023.
\newblock hal-04019749.

\bibitem[GLR{\etalchar{+}}22]{goldt2022gaussian}
Sebastian Goldt, Bruno Loureiro, Galen Reeves, Florent Krzakala, Marc
  M{\'e}zard, and Lenka Zdeborov{\'a}.
\newblock The gaussian equivalence of generative models for learning with
  shallow neural networks.
\newblock In {\em Mathematical and Scientific Machine Learning}, pages
  426--471. PMLR, 2022.

\bibitem[GN98]{gray1998quantization}
Robert~M. Gray and David~L. Neuhoff.
\newblock Quantization.
\newblock {\em IEEE transactions on information theory}, 44(6):2325--2383,
  1998.

\bibitem[HMD15]{han2015deep}
Song Han, Huizi Mao, and William~J Dally.
\newblock Deep compression: Compressing deep neural networks with pruning,
  trained quantization and huffman coding.
\newblock {\em arXiv preprint arXiv:1510.00149}, 2015.

\bibitem[HMRT22]{hastie2022surprises}
Trevor Hastie, Andrea Montanari, Saharon Rosset, and Ryan~J Tibshirani.
\newblock Surprises in high-dimensional ridgeless least squares interpolation.
\newblock {\em Annals of statistics}, 50(2):949, 2022.

\bibitem[HTL{\etalchar{+}}23]{hernandez2023training}
Charles Hernandez, Bijan Taslimi, Hung~Yi Lee, Hongcheng Liu, and Panos~M
  Pardalos.
\newblock Training generalizable quantized deep neural nets.
\newblock {\em Expert Systems with Applications}, 213:118736, 2023.

\bibitem[Hua17]{huang2017statistical}
Haiping Huang.
\newblock Statistical mechanics of unsupervised feature learning in a
  restricted boltzmann machine with binary synapses.
\newblock {\em Journal of Statistical Mechanics: Theory and Experiment},
  2017(5):053302, 2017.

\bibitem[KH92]{Krogh1992}
Anders Krogh and John~A Hertz.
\newblock Generalization in a linear perceptron in the presence of noise.
\newblock {\em Journal of Physics A: Mathematical and General}, 25(5):1135,
  1992.

\bibitem[KMS{\etalchar{+}}12a]{Krzakala2012PRX}
F~Krzakala, M~M\'{e}zard, F~Sausset, Y~F.~Sun, and L~Zdeborov\'{a}.
\newblock Statistical-physics-based reconstruction in compressed sensing.
\newblock {\em Physical Review X}, 2:021005, 2012.

\bibitem[KMS{\etalchar{+}}12b]{Krzakala2012JSTAT}
Florent Krzakala, Marc M{\'e}zard, Francois Sausset, Yifan Sun, and Lenka
  Zdeborov{\'a}.
\newblock Probabilistic reconstruction in compressed sensing: algorithms, phase
  diagrams, and threshold achieving matrices.
\newblock {\em Journal of Statistical Mechanics: Theory and Experiment},
  2012(08):P08009, 2012.

\bibitem[KWT09]{kabashima2009typical}
Yoshiyuki Kabashima, Tadashi Wadayama, and Toshiyuki Tanaka.
\newblock A typical reconstruction limit for compressed sensing based on
  lp-norm minimization.
\newblock {\em Journal of Statistical Mechanics: Theory and Experiment},
  2009(09):L09003, 2009.

\bibitem[LDX{\etalchar{+}}17]{li2017training}
Hao Li, Soham De, Zheng Xu, Christoph Studer, Hanan Samet, and Tom Goldstein.
\newblock Training quantized nets: A deeper understanding.
\newblock {\em Advances in Neural Information Processing Systems}, 30, 2017.

\bibitem[LGC{\etalchar{+}}21]{loureiro2021learning}
Bruno Loureiro, Cedric Gerbelot, Hugo Cui, Sebastian Goldt, Florent Krzakala,
  Marc Mezard, and Lenka Zdeborov{\'a}.
\newblock Learning curves of generic features maps for realistic datasets with
  a teacher-student model.
\newblock {\em Advances in Neural Information Processing Systems},
  34:18137--18151, 2021.

\bibitem[LGW{\etalchar{+}}21]{liang2021pruning}
Tailin Liang, John Glossner, Lei Wang, Shaobo Shi, and Xiaotong Zhang.
\newblock Pruning and quantization for deep neural network acceleration: A
  survey.
\newblock {\em Neurocomputing}, 461:370--403, 2021.

\bibitem[LVM{\etalchar{+}}20]{loog2020brief}
Marco Loog, Tom Viering, Alexander Mey, Jesse~H Krijthe, and David~MJ Tax.
\newblock A brief prehistory of double descent.
\newblock {\em Proceedings of the National Academy of Sciences},
  117(20):10625--10626, 2020.

\bibitem[MLM16]{miyashita2016convolutional}
Daisuke Miyashita, Edward~H Lee, and Boris Murmann.
\newblock Convolutional neural networks using logarithmic data representation.
\newblock {\em arXiv preprint arXiv:1603.01025}, 2016.

\bibitem[MPV87]{Mezard1987}
Marc M{\'e}zard, Giorgio Parisi, and Miguel Virasoro.
\newblock {\em Spin glass theory and beyond: An Introduction to the Replica
  Method and Its Applications}, volume~9.
\newblock World Scientific Publishing Co Inc, 1987.

\bibitem[MS22]{montanari2022universality}
Andrea Montanari and Basil~N Saeed.
\newblock Universality of empirical risk minimization.
\newblock In {\em Conference on Learning Theory}, pages 4310--4312. PMLR, 2022.

\bibitem[Ran11]{Rangan2011}
Sundeep Rangan.
\newblock Generalized approximate message passing for estimation with random
  linear mixing.
\newblock In {\em Information Theory Proceedings (ISIT), 2011 IEEE
  International Symposium on}, pages 2168--2172. IEEE, 2011.

\bibitem[RTWZ01]{ricci2001simplest}
Federico Ricci-Tersenghi, Martin Weigt, and Riccardo Zecchina.
\newblock Simplest random k-satisfiability problem.
\newblock {\em Physical Review E}, 63(2):026702, 2001.

\bibitem[SA21]{sasaki2021analysis}
Ryuta Sasaki and Toru Aonishi.
\newblock Analysis of the hopfield model with discrete coupling.
\newblock {\em Journal of the Physical Society of Japan}, 90(9):094602, 2021.

\bibitem[Sak23]{sakata2023prediction}
Ayaka Sakata.
\newblock Prediction errors for penalized regressions based on generalized
  approximate message passing.
\newblock {\em Journal of Physics A: Mathematical and Theoretical},
  56(4):043001, 2023.

\bibitem[SO21]{sakata2021perfect}
Ayaka Sakata and Tomoyuki Obuchi.
\newblock Perfect reconstruction of sparse signals with piecewise continuous
  nonconvex penalties and nonconvexity control.
\newblock {\em Journal of Statistical Mechanics: Theory and Experiment},
  2021(9):093401, 2021.

\bibitem[WCB{\etalchar{+}}18]{wang2018training}
Naigang Wang, Jungwook Choi, Daniel Brand, Chia-Yu Chen, and Kailash
  Gopalakrishnan.
\newblock Training deep neural networks with 8-bit floating point numbers.
\newblock {\em Advances in neural information processing systems}, 31, 2018.

\bibitem[WJZ{\etalchar{+}}20]{wu2020integer}
Hao Wu, Patrick Judd, Xiaojie Zhang, Mikhail Isaev, and Paulius Micikevicius.
\newblock Integer quantization for deep learning inference: Principles and
  empirical evaluation.
\newblock {\em arXiv preprint arXiv:2004.09602}, 2020.

\bibitem[ZHMD16]{zhu2016trained}
Chenzhuo Zhu, Song Han, Huizi Mao, and William~J Dally.
\newblock Trained ternary quantization.
\newblock In {\em International Conference on Learning Representations}, 2016.

\bibitem[ZHT07]{zou2007degree}
H.~Zou, T.~Hastie, and R.~Tibshirani.
\newblock On the degrees of freedom of the lasso.
\newblock {\em Annal. Stat.}, 35(5):2173--2192, 2007.

\bibitem[ZK16]{Zdeborova2016}
Lenka Zdeborov{\'a} and Florent Krzakala.
\newblock Statistical physics of inference: Thresholds and algorithms.
\newblock {\em Advances in Physics}, 65(5):453--552, 2016.

\bibitem[ZYG{\etalchar{+}}16]{zhou2016incremental}
Aojun Zhou, Anbang Yao, Yiwen Guo, Lin Xu, and Yurong Chen.
\newblock Incremental network quantization: Towards lossless cnns with
  low-precision weights.
\newblock In {\em International Conference on Learning Representations}, 2016.

\end{thebibliography}
\appendix
\section{Additional Result}
\begin{figure}[h]
    \begin{centering}
  \includegraphics[width = 140mm]{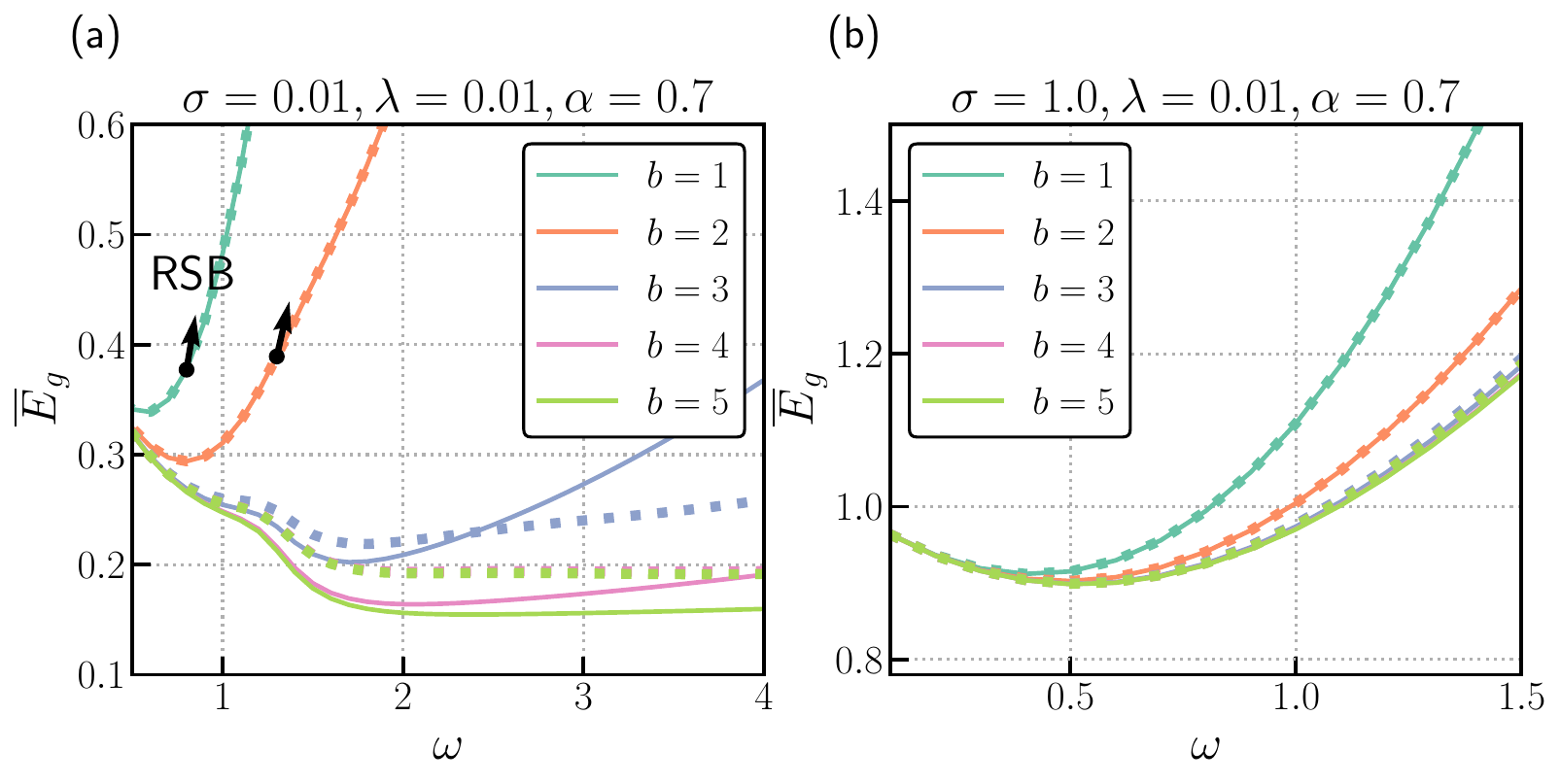}
      \end{centering}
      \caption{
     Expected generalization error under RS assumption as a function of $\omega$ for (a) $(\sigma, \lambda, \alpha)=(0.01, 0.01, 0.7)$ and (b) $(\sigma, \lambda, \alpha)=(1.0, 0.01, 0.7)$.
The outcome of uniform quantization is represented by solid lines, while non-uniform quantization is represented by dashed lines. The black arrows indicate the RSB phase.
      }
      \label{appe_1}
\end{figure}

We present an additional finding regarding the analysis of the expected generalization error. Figure \ref{appe_1} shows the generalization errors for both the low noise case (a) and the high noise case (b) when $\lambda = 0.01$ and $\alpha = 0.7$. It is evident that the generalization performance is significantly worse due to the impact of overparameterization, where $\alpha < 1$, compared to the performance shown in Figure 4 of the main text.

\begin{figure}[h]
    \begin{centering}
  \includegraphics[width = 140mm]{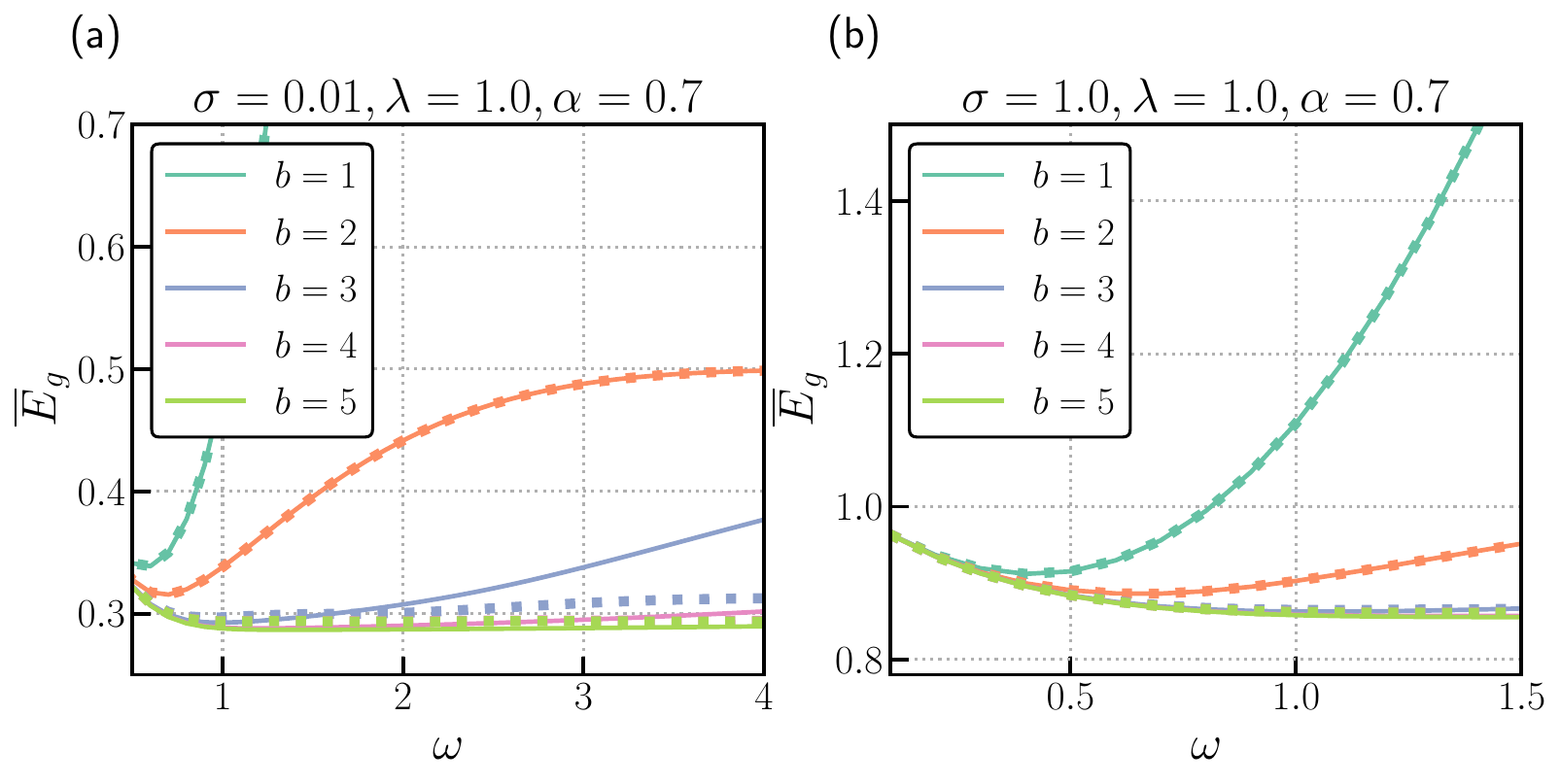}
      \end{centering}
      \caption{
     Expected generalization error under RS assumption as a function of $\omega$ for (a) $(\sigma, \lambda, \alpha)=(0.01, 1.0, 0.7)$ and (b) $(\sigma, \lambda, \alpha)=(1.0, 1.0, 0.7)$.
Solid lines and dashed lines represent the result of uniform and non-uniform quantization, respectively. RSB phase is indicated by black arrows. 
      }
      \label{appe_2}
\end{figure}

Figure \ref{appe_2} shows the generalization errors with (a) low and (b) high noise cases at $\lambda = 1.0$ and $\alpha = 0.7$.
We observe that regularization can mitigate the sharpness of the optimality condition for $\omega$.

\begin{figure}[h]
    \begin{centering}
  \includegraphics[width = 140mm]{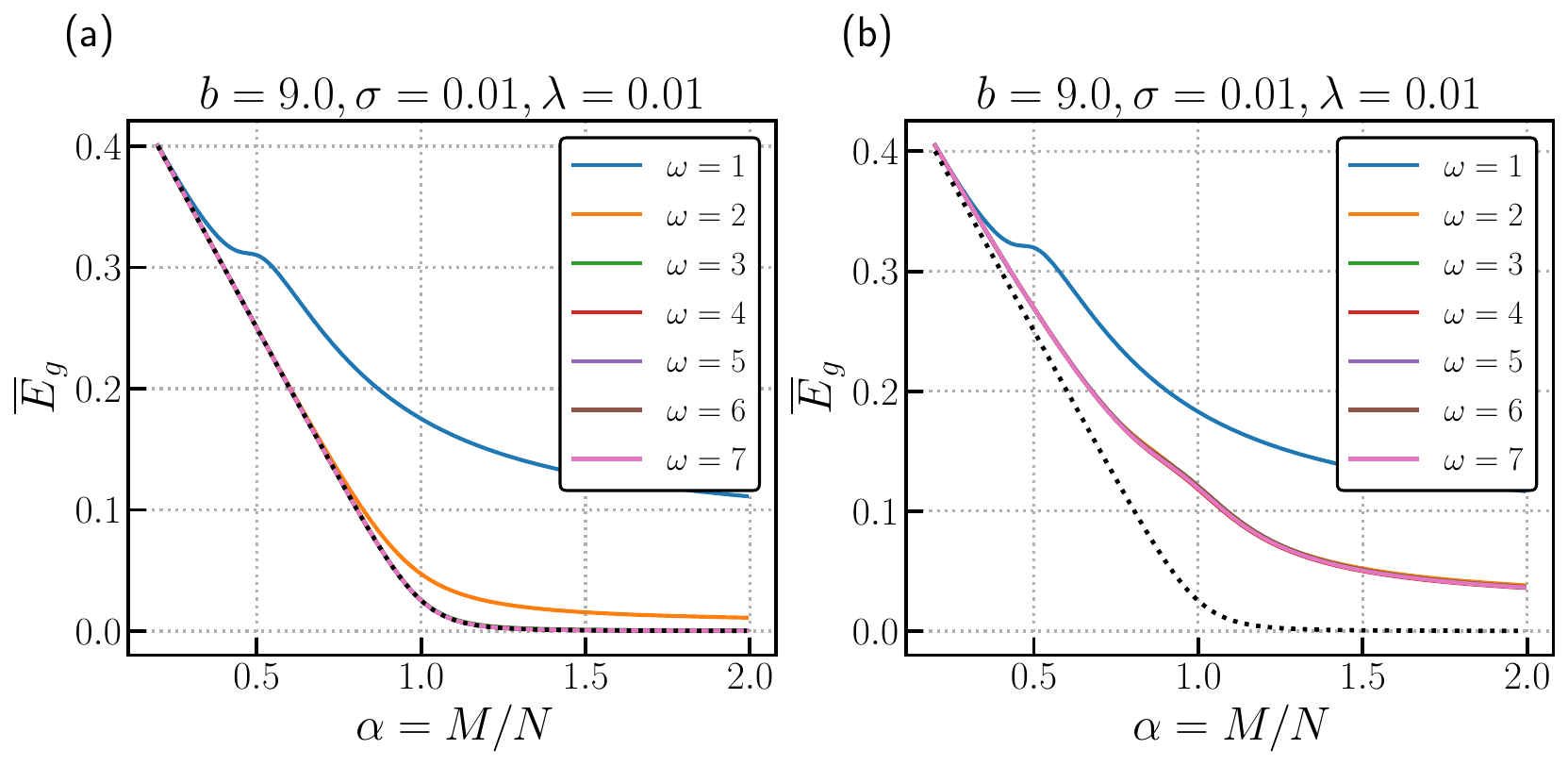}
      \end{centering}
      \caption{
      Expected generalization error under RS assumption as a function of $\alpha = N/M$ for (a) uniform quantization and (b) non-uniform quantization.
The dashed lines represent the result of ridge regression.}
      \label{appe_3}
\end{figure}

Figure \ref{appe_3} shows
the generalization errors under quantization as a function of $\alpha$
for (a) uniform and (b) non-uniform cases.
The double decent peak is diminished in the low-noise case.

\begin{figure}[h]
    \begin{centering}
  \includegraphics[width = 140mm]{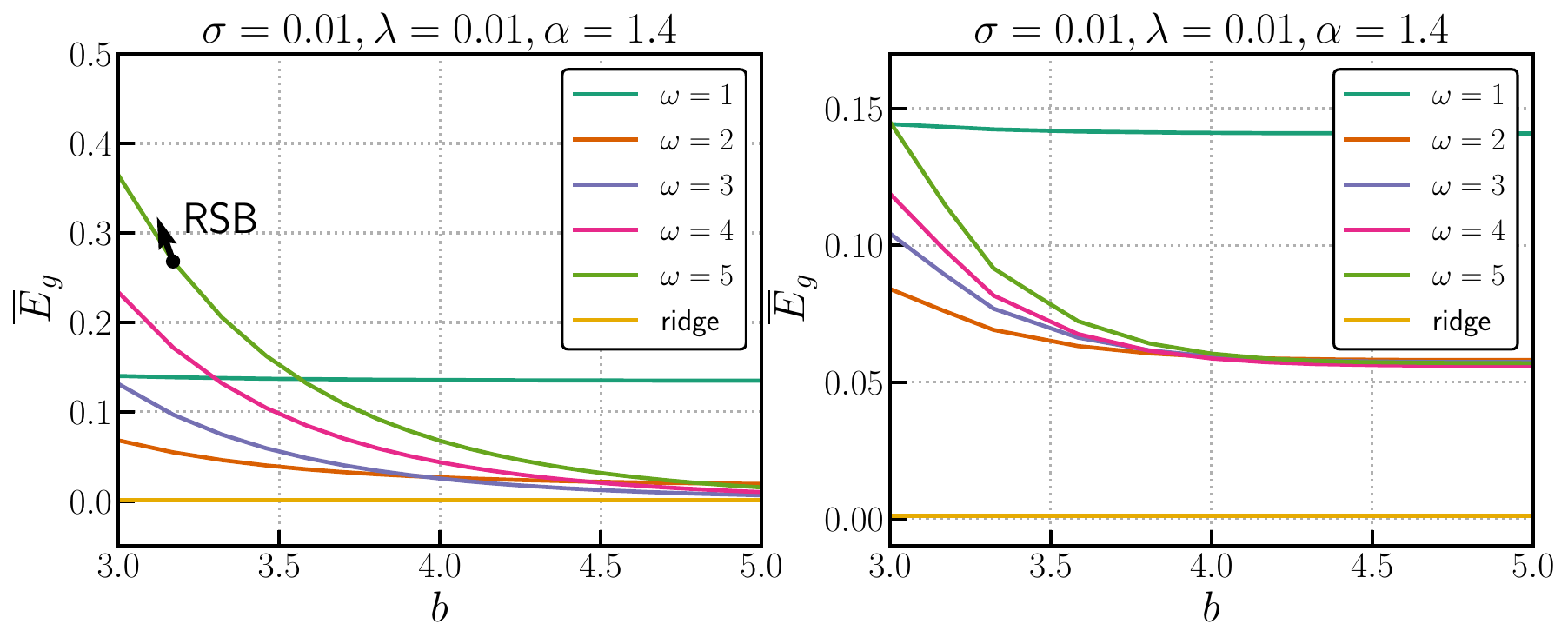}
      \end{centering}
      \caption{
      The expected generalization error under the RS assumption is presented as a function of $b$ for (a) uniform quantization and (b) non-uniform quantization. The yellow lines represent the results of the ridge regression.}
      \label{appe_4}
\end{figure}

Figure \ref{appe_4} illustrates the generalization errors associated with quantization as a function of the number of bits $b$ for (a) uniform and (b) non-uniform cases.
In the case of uniform quantization, the generalization performance asymptotically approaches that of the ridge regression as the number of bits increases.
This is not the case, however, for non-uniform quantization.
In non-uniform quantization, since the partition width increases exponentially, the approximation accuracy is inferior compared to that of uniform quantization.

\section{State Evolution and Local Stability}
The typical performance of AMP can be analyzed by state evolution (SE).
SE describes the statistical properties of the trajectory of AMP, which depend on the randomness of the data.
The typical trajectory of AMP is tracked by two variables: $V^t := \frac{1}{N}\sum_i^N v_i^t$ and $E^t := \frac{1}{N}\sum_i^N (w_i^0-\bar{m}_i^t)$.
Assuming the same settings as the replica analysis in the main text, the SE equations are given as
\begin{align}
V^{t+1} &= \int Dz \frac{\partial \varphi^*\left(\xi^t z, \Lambda^t\right)}{\partial( \xi^t z)},\\
E^{t+1} &= \rho - 2\frac{\rho \alpha}{\xi^t (1+V^t)}\int Dz \frac{\partial \varphi^* (\xi^t z, \Lambda^t)}{\partial z}+\int Dz \varphi^*(\xi^t z, \Lambda^t)^2,
\end{align}
with $\xi^t := \sqrt{\alpha^2 \rho/(1+V^t)^2 + \alpha (\sigma^2 + E^t)/(1+V^t)^2 }$ and $\Lambda^t:=\lambda + \alpha/(1+V^t)$.
The SE equations and the saddle point equations of the replica analysis in the main text are equivalent based on the correspondences
\begin{align}
V^t &\leftrightarrow \chi, \\
E^t &\leftrightarrow Q-2m +\rho.
\end{align}
Therefore, the fixed point of SE and the RS solution of the replica analysis are equivalent.
Furthermore, the linear stability condition of the fixed point of AMP is derived as
\begin{align}
\frac{\alpha}{(1+V)^2} \int Dz \left(\partial \varphi^*(\xi z, \Lambda)\right)^2 < 1.
\end{align}
Since this condition is equivalent to the stability condition of the RS solution in replica analysis, it is expected that theoretical results obtained by the replica method in the RS phase can be validated through numerical experiments employing AMP.

\end{document}